\documentclass[sigconf]{acmart}
\usepackage{fancyhdr}
\usepackage{balance}
\usepackage{setspace}
\usepackage{multirow}
\usepackage{booktabs} 
\usepackage{enumitem} 
\usepackage{threeparttable} 
\usepackage{graphicx} 
\usepackage{subfigure} 
\usepackage{enumerate}
\usepackage{bm}
\newsavebox\CBox
\def\textBF#1{\sbox\CBox{#1}\resizebox{\wd\CBox}{\ht\CBox}{\textbf{#1}}}
\AtBeginDocument{%
  }

\copyrightyear{2022}
\acmYear{2022}
\setcopyright{acmcopyright}\acmConference[SIGIR '22]{Proceedings of the 45th
International ACM SIGIR Conference on Research and Development in Information
Retrieval}{July 11--15, 2022}{Madrid, Spain}
\acmBooktitle{Proceedings of the 45th International ACM SIGIR Conference on
Research and Development in Information Retrieval (SIGIR '22), July 11--15, 2022,
Madrid, Spain}
\acmPrice{15.00}
\acmDOI{10.1145/3477495.3532011}
\acmISBN{978-1-4503-8732-3/22/07}

\begin{document}
\fancyhead{}

\title{Learning Disentangled Representations for Counterfactual Regression via Mutual Information Minimization}
\author{Mingyuan Cheng}
\email{wanyu.cmy@alibaba-inc.com}
\orcid{0000-0001-5159-0299}
\affiliation{%
  \institution{Alibaba Group}
  \city{Beijing}
  \country{China}
}
\author{Xinru Liao}
\email{xinru.lxr@alibaba-inc.com}
\affiliation{%
  \institution{Alibaba Group}
  \city{Hangzhou}
  \country{China}
}
\author{Quan Liu}
\authornote{Corresponding author.}
\email{lq204691@alibaba-inc.com}
\affiliation{%
  \institution{Alibaba Group}
  \city{Hangzhou}
  \country{China}
}
\author{Bin Ma}
\email{mabin.mb@alibaba-inc.com}
\affiliation{%
  \institution{Alibaba Group}
  \city{Hangzhou}
  \country{China}
}
\author{Jian Xu}
\email{xiyu.xj@alibaba-inc.com}
\affiliation{%
  \institution{Alibaba Group}
  \city{Beijing}
  \country{China}
}

\author{Bo Zheng}
\email{bozheng@alibaba-inc.com}
\affiliation{%
  \institution{Alibaba Group}
  \city{Beijing}
  \country{China}
}
\renewcommand{\shortauthors}{Cheng and Liao, et al.}

\begin{abstract}
Learning individual-level treatment effect is a fundamental problem in causal inference and has received increasing attention in many areas, especially in the user growth area which concerns many internet companies. Recently, disentangled representation learning methods that decompose covariates into three latent factors, including instrumental, confounding and adjustment factors, have witnessed great success in treatment effect estimation. However, it remains an open problem how to learn the underlying disentangled factors precisely. Specifically, previous methods fail to obtain independent disentangled factors, which is a necessary condition for identifying treatment effect. In this paper, we propose Disentangled Representations for Counterfactual Regression via Mutual Information Minimization (MIM-DRCFR), which uses a multi-task learning framework to share information when learning the latent factors and incorporates MI minimization learning criteria to ensure the independence of these factors. Extensive experiments including public benchmarks and real-world industrial user growth datasets demonstrate that our method performs much better than state-of-the-art methods.
\end{abstract}



\maketitle

\section{Introduction}
Estimating treatment effect is one of the most important topics in many domains, such as policy making \cite{lalonde1986evaluating,Athey2016RecursivePF}, medicine prediction~\citep{Shalit2017EstimatingIT}, advertisement~\citep{Bottou2013CounterfactualRA,Sun2015CausalIV,gu2021estimating}, recommendation \citep{zhang2021causal,wang2021counterfactual} and user growth \cite{du2019improve}. It often needs to answer counterfactual problems~\citep{pearl2009causality} like \textit{“Would this patient have low blood sugar had she received a medication?”} or \textit{“Would the customer buy the product had he got a 70\% discount?”}. 
Specifically, in the user growth area, companies may take many activities such as sending coupons and pushing messages to increase user acquisition or retention, where the counterfactual problem becomes
\textit{“Would the user act more actively on the platform had he received the coupon or message?”.}\par
One golden standard approach to learn causal effect is to perform Randomized Controlled Trial~\citep{pearl2009causality}, where the treatment is randomly assigned to individuals. However, this is often expensive, unethical or even infeasible, thus we usually focus on estimating treatment effect from observational data. In the observational study, the treatment often depends on some attributes of the individual $x$, which causes the problem of \textbf{selection bias}~\citep{Imbens2015CausalIF} (i.e., $p(t|x)\neq p(t)$). Taking the medicine scenario for example, the economic status affects both the medications and the patient's recovery rate. And it is vital to find all such confounding variables (i.e., affecting both the treatment and outcome) and control them to make precise predictions. This means \textit{unconfoundedness} assumption often needs to be satisfied in the observational study to make the treatment effect identifiable~\citep{pearl2009causality}.\par
Even though we already have all confounders in our variables, we still face a difficult problem of identifying them and further balancing them with the \textit{backdoor criterion} ~\citep{pearl2009causality}. Existing methods achieve balancing either by propensity score weighting methods~\citep{Austin2011AnIT} or representation learning methods which reduce the discrepancy between the treated and control group (e.g., BNN~\citep{Johansson2016LearningRF} and CFR-net~\citep{Shalit2017EstimatingIT}) while ignoring identification of other latent factors. Recently, disentangled representation learning methods, $\mathrm{D}^{2}$VD~\citep{Kuang2017TreatmentEE}, DR-CFR~\citep{Hassanpour2020LearningDR} and TEDVAE~\citep{Zhang2021TreatmentEE} have been proposed to learn disentangled factors $\left \{ \Gamma,\Upsilon,\Delta \right \}$, respectively representing factor that affects only the treatment, only the outcome, and both the treatment and the outcome (\textit{aka} instrumental, adjustment and confounding factors). Although disentangled representation learning methods greatly help achieve exact identification of the latent factors, the above methods still face the following limitations: $\mathrm{D}^{2}$VD only decomposes features into two factors $\left \{\Upsilon,\Delta \right \}$, DR-CFR cannot effectively distinguish the difference between the $\Delta$ and $\left \{ \Gamma,\Upsilon \right \}$ and TEDVAE uses the generative model which might greatly increase model training complexity. Besides, these methods cannot obtain \textbf{independent disentangled representations}, which is a necessary condition for identifying treatment effect. To solve the problem of independence of disentangled representations, we propose to use MI minimization \citep{Cheng2020CLUBAC,Chen2018IsolatingSO} method, which has obtained increasing attention in domain adaptation \cite{granger1994using} and style transfer \cite{kazemi2018unsupervised} recently. It is typically utilized as a learning criterion in loss function to ensure the independence between variables. Specifically, \citep{Cheng2020CLUBAC} proposes a MI upper bound called Contrastive Log-Ratio Upper Bound (CLUB) to deal with the MI minimization task and various experiments have demonstrated the effectiveness of this method. \par
In this paper, we propose an easy-handling and well-identifying model to deal with the problems of disentanglement in treatment effect estimation. We incorporate the multi-task learning framework such as shared-bottom structure and MI minimization criteria to learn the disentangled factors. And our main contributions are:
\begin{itemize}
\item We propose a multi-task learning structure represented by the disentangled representation layer to share information across these latent factors, instead of three independent representation networks which is commonly used by previous work.
\item We introduce the MI minimization method into causal inference to learn the latent factors, which uses CLUB as MI upper bound to obtain ideally independent disentangled representations.
\item We carry out extensive experiments on both public benchmarks and industrial datasets of user growth (e.g., message pushing and coupon sending), which demonstrate the superiority of our method.
\end{itemize}

\section{The Proposed Method}
\begin{figure}[]
\centerline{\includegraphics[width=0.45\textwidth,height=0.35\textwidth]{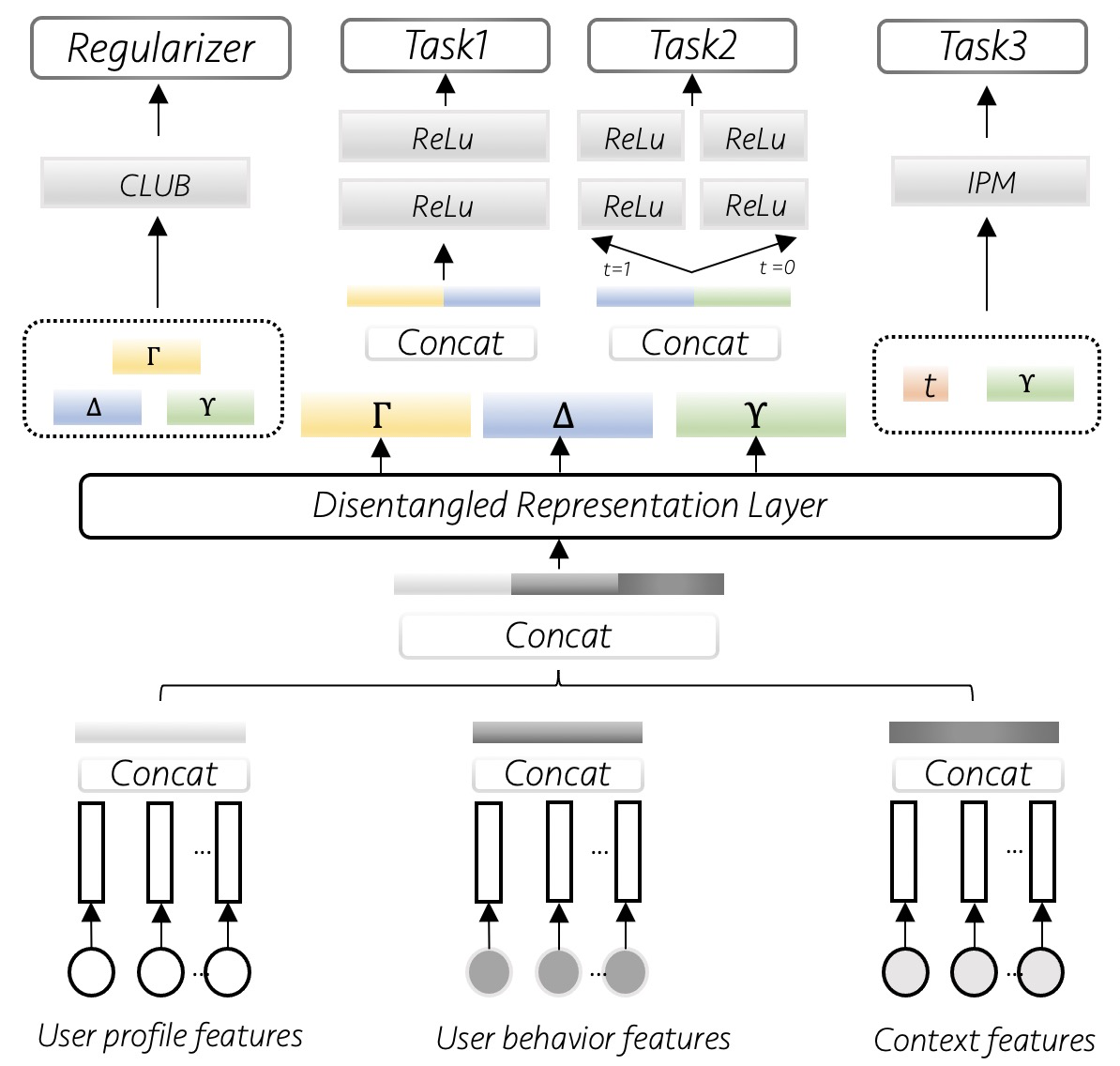}}
\hypertarget{model}{\caption{The proposed model architecture of MIM-DRCFR in our industrial datasets. The input variable contains users’ profile, behavior and context features. The disentangled representation layer consists of a shared-bottom structure and three factor-specific layers and then outputs three latent factors $\left \{ \Gamma,\Delta,\Upsilon \right \}$. The top of the figure shows multi-task objectives, including treatment prediction, outcome prediction, IPM constraint and the MI minimization regularizer.}}
\label{fig:model}
\end{figure}

In this section, we introduce our model architecture as shown in Figure \ref{fig:model}. We first present the basic definition and assumption in 2.1. Then, we explain the multi-task disentangled representations learning framework in 2.2. Last, we illustrate MI minimization regularizer for causal inference in 2.3.
\subsection{Preliminary}
We first present some notations in our context. Given the observational dataset $\mathcal{D}= \{ ( x_{i}, t_{i}, y_{i}^{t_{i}}(x_{i},t_{i}) )  \}_{i=1}^{n}$, where $n$ is the number of data samples, 
$x_{i} \in \mathcal{X}$ is the input features referring individual context information, $y_{i}^{t_{i}}(x_{i},t_{i}) \in \mathcal{Y}$ is observed factual outcome and counterfactual outcome $y_{i}^{1-t_{i}}(x_{i},t_{i})$ is missing here, $t_{i} \in \mathcal{T}$ refers to potential interventions (e.g., for binary treatment  $t \in \left \{0,1\right \}$). Mathematically, we define our goal in this paper is to learn a function $\mathcal{F}: \mathcal{X} \times \mathcal{T} \to \mathcal{Y}$ to predict the potential outcomes and then estimate the \textit{individual treatment effect (ITE)}\footnote{The individual treatment effect (ITE), \textit{aka} conditional average treatment effect (CATE).} and the \textit{average treatment effect (ATE):}\par
\textbf{Definition 1.} \textit{The individual treatment effect is formulated as}:
\begin{equation}
    \tau_{i}=y_{i}^{1}(x_{i},t_{i})-y_{i}^{0}(x_{i},t_{i})
\end{equation}\par
\textbf{Definition 2.} \textit{The average treatment effect is formulated as}:
\begin{equation}
    ATE=\frac{1}{n}\sum_{i=1}^{n}\tau_{i}
\end{equation}\par
The following fundamental assumptions ~\citep{Rosenbaum1983TheCR} need to be satisfied in individual treatment effect estimation:\par
\textbf{Assumption 1}. (\textit{\textbf{SUTVA}}) \textit{The Stable Unit Treatment Value Assumption requires that the response of a unit depends only on the treatment to which he himself was assigned and not affected by others.}
\par \textbf{Assumption 2}. (\textit{\textbf{Unconfoundedness}}) \textit{The treatment assignment mechanism is independent of the potential outcome when conditioning on the observed variables, Formally}: $Y_{0}, Y_{1} \perp \!\!\!\! \perp t \mid x$.\par
\textbf{Assumption 3}. (\textit{\textbf{Positivity}}) \textit{Each unit has a non-zero probability to be assigned to each treatment when given the observed contexts, i.e.,} $0 < P(t = 1|x) < 1.$
\subsection{Multi-task Disentangled Representation Learning}
Without loss of generality, we assume the dataset $\mathcal{D}$ is generated from three underlying factors $\left \{ \Gamma,\Delta,\Upsilon \right \}$~\citep{Hassanpour2020LearningDR}. In our user growth scene, $x_{i}$ consists of user’s profiles, behavior and context features. $t_{i}$ can be sending coupon or pushing message to user. $y_{i}$ can be user’s login rate or click-through rate. Then we aim to encode the input features $x_{i}$ into three separate embedding parts through disentangled representation layer ($DRL$), which is formulated as:
\begin{equation}
\Gamma,\Delta,\Upsilon = DRL(x),
\end{equation}
where $x \subseteq \mathbb{R}^{1 \times d}$ and $\Gamma,\Delta,\Upsilon \subseteq \mathbb{R}^{1 \times d}$, $d$ refers to feature dim. \par
DR-CFR directly uses three separate representation networks to learn these factors, while our experiments show that this method cannot effectively distinguish the difference between the $\Delta$ and $\left \{ \Gamma,\Upsilon \right \}$. Inspired by multi-task learning, we use a shared-bottom structure to learn the feature embedding from input variables and then use three factor-specific layers to decode the embedding into latent factors $\left \{ \Gamma,\Delta,\Upsilon \right \}$ (\textit{aka} SFD layer\footnote{named from \textbf{S}hared-bottom and \textbf{F}actor-specific \textbf{D}isentangled representation layer.}). Then we learn the latent factors by following tasks:\par
\textbf{Task1.} Predict the treatment from $\Omega=\textsc{concat}(\Gamma,\Delta)$ and define the loss $\mathcal{L}_{\text{treat}}=\mathcal{L}[t_{i}, \pi(\Omega(x_{i}))]$. $\pi$ is a classifier. Minimizing the loss $\mathcal{L}_{\text{treat}}$ ensures that the information of treatment is captured in the union of $\Gamma$ and $\Delta$.\par
\textbf{Task2.} Predict the outcome from $\Phi=\textsc{concat}(\Upsilon,\Delta)$ and define the loss $\mathcal{L}_{\text{pred}}=\mathcal{L}[y_{i}, h^{t_{i}}(\Phi(x_{i}))]$. $h^{t_{i}}$ is regression network for each treatment arm. We ensure the information of outcome is captured in the union of $\Upsilon$ and $\Delta$ by minimizing $\mathcal{L}_{\text{pred}}$.\par
\textbf{Task3.} Restrict discrepancy distance and define the loss as $\mathcal{L}_{\text{disc}}=\textrm{IPM} \left (\left \{ \Upsilon(x_{i}) \right \}_{i:t_{i}=0}, \left \{ \Upsilon(x_{i}) \right \}_{i:t_{i}=1}  \right )$. We ensure the latent factor $\Upsilon$ is irrelevant to treatment by minimizing $\mathcal{L}_{\text{disc}}$. \par
We expect that all confounding factors are captured in $\Delta$ when we can completely distinguish $\Delta$ from $\left \{ \Upsilon,\Gamma\right \}$ by the following multi-objective function:
\begin{equation}
\label{E1}
    \mathcal{L}_{\textsc{main}}=\mathcal{L}_{\text{pred}} + \alpha \cdot \mathcal{L}_{\text{treat}} +\beta \cdot \mathcal{L}_{\text{disc}}
\end{equation}
where $\alpha$ and $\beta$ are weights for each task, and we use \textit{Wasserstein distance} as our integral probability metric in this paper.\par

\subsection{MI Minimization Regularizer}
To obtain independent disentangled factors, we propose to minimize the MI among the three factors to ensure independence. MI is a fundamental measure of the dependence between two random variables. Mathematically, the definition of MI between variables $x$ and $y$ is:
\begin{equation}
\label{E1}
    \mathrm{I}(x,y)=\mathbb{E}_{p(x,y)}\left [ \log \frac{p(x,y)}{p(x)p(y)}\right]
\end{equation}\par
Following ~\citep{Cheng2020CLUBAC}, we introduce using CLUB as MI upper bound to accomplish MI minimization among latent factors, and \textsc{CLUB} is defined as $\mathrm{I}_{\textsc{club}}(x,y)=\mathbb{E}_{p(x,y)}\left [\log p(y|x) \right ] - \mathbb{E}_{p(x)}\mathbb{E}_{p(y)}\left [\log p(y|x) \right ]$ when the conditional distribution $p(y|x)$ is known. Unfortunately, the conditional relations between variables are unavailable in our task, and therefore we use a variational distribution $q_{\theta}(y|x)$ to approximate $p(y|x) $ and further extend the CLUB estimator into vCLUB, which is defined as $\mathrm{I}_{\mathrm{v}\textsc{club}}(x,y)=\mathbb{E}_{p(x,y)}\left [\log q_{\theta}(y|x) \right ] - \mathbb{E}_{p(x)}\mathbb{E}_{p(y)}\left [\log q_{\theta}(y|x) \right ]$. $\mathrm{I}_{\mathrm{v}\textsc{club}}(x,y)$ remains a MI upper bound when we have good variational approximation $q_{\theta}(y|x)$.\par
Then we use the following MI minimization regularizer to obtain independent disentangled representations for ITE estimation:
\begin{equation}
\label{E1}
    \mathcal{L}_{\textsc{club}} = \mathrm{I}_{\mathrm{v}\textsc{club}}(\Gamma,\Delta) + \mathrm{I}_{\mathrm{v}\textsc{club}}(\Delta,\Upsilon) + \mathrm{I}_{\mathrm{v}\textsc{club}}(\Upsilon,\Gamma)
\end{equation}\par
We summarize the total objective function $\mathcal{L_{\textit{MIM-DRCFR}}}$ as:
\begin{equation}
\label{E1}
    \mathcal{L_{\textit{MIM-DRCFR}}} = \mathcal{L}_{\text{pred}} + \alpha \cdot \mathcal{L}_{\text{treat}} +\beta \cdot \mathcal{L}_{\text{disc}} + \gamma \cdot \mathcal{L}_{\textsc{club}} +\lambda \cdot \mathcal{L}_{\textsc{reg}}
\end{equation}
where $\mathcal{L}_{\textsc{reg}}$ is used to penalize the model complexity and $\alpha$, $\beta$, $\gamma$ and $\lambda$ are weights for these objectives.\par
Besides, inspired by the orthogonal regularizer in $\mathrm{D}^{2}$VD~\citep{Kuang2017TreatmentEE}, we introduce a criterion called \textbf{R}epresentation \textbf{L}ayer \textbf{O}rthogonality (\textsc{RLO}), which is an intuitive method to obtain disjoint factors:
\begin{equation}
\label{E1}
    \mathcal{L}_{\textsc{rlo}} = \bar{W}_{\Gamma}^{T} \cdot \bar{W}_{\Delta} + \bar{W}_{\Delta}^{T} \cdot \bar{W}_{\Upsilon} + \bar{W}_{\Upsilon}^{T} \cdot \bar{W}_{\Gamma}
\end{equation}
where $W \subseteq \mathbb{R}^{d \times d}$ refers to products of the $DRL$, then we calculate average of $W$ ($\bar{W} \subseteq \mathbb{R}^{d \times 1}$) to represent the contribution of input variables on disentangled factors. By minimizing  $\mathcal{L}_{\textsc{rlo}}$, we expect each dimension of $x$ is only embedded in one of $\left \{ \Gamma,\Delta,\Upsilon \right \}$. $\mathcal{L_{\textit{RLO-DRCFR}}}$ is obtained by replacing $\mathcal{L}_{\textsc{club}}$ with $\mathcal{L}_{\textsc{rlo}}$. 

\section{Experiment}
\subsection{Benchmark Evaluation}
A fundamental problem in causal inference is that we cannot observe factual outcome and counterfactual outcome simultaneously. One common used solution is to synthesize datasets where the outcomes of all possible treatments are available or synthesize outcomes from real-world covariates.\par
\textbf{IHDP Benchmark}. Similar to~\citep{Shalit2017EstimatingIT,Hassanpour2019CounterFactualRW,Zhang2021TreatmentEE}, we use a semi-synthetic dataset based on IHDP as our benchmark which was first introduced by~\citep{Hill2011BayesianNM}. The covariates come from a randomized experiment studying the effects of home visits by specialist on future cognitive test scores. The selection bias was introduced by removing a biased subset of the treated population and it comprises 747 instances (139 treated, 608 control) with 25 covariates measuring different attributes of children and their mothers. The simulated outcomes are implemented as both setting “A” and setting “B” in the NPCI package and follow \textit{linear} and \textit{nonlinear} relationship respectively.\par

\textbf{Performance Metrics}. Given a synthetic dataset that includes both factual and counterfactual outcomes, we evaluate treatment effect estimation methods through two performance measures. The individual-based metric is $\epsilon_{\textit{PEHE}}=\frac{1}{n}\sum_{i=1}^{n}{(\hat{\tau}_{i}-\tau_{i})^{2}}$, where $\hat{\tau}_{i}=\hat{y}_{i}^{1}-\hat{y}_{i}^{0}$ is the predicted individual treatment effect and $\tau_{i}=y_{i}^{1}-y_{i}^{0}$ is the actual effect.
The population-based measure is $\epsilon_{\textit{ATE}}=\vert \mathrm{ATE}-\widehat{\mathrm{ATE}}\vert$. ATE $=\frac{1}{n}\sum_{i=1}^{n}(y_{i}^{1}-y_{i}^{0})$ and $\widehat{\mathrm{ATE}}$ is calculated from the estimated outcomes.\par 

\textbf{Baselines Methods}. We compare performances of the following methods which can be divided into: \textit{Baseline models}: \textbf{TARNET}~\citep{Shalit2017EstimatingIT}, \textbf{CFR-WASS}~\citep{Shalit2017EstimatingIT}, \textbf{CFR-MMD}~\citep{Shalit2017EstimatingIT}, \textbf{CFR-ISW}~\citep{Hassanpour2019CounterFactualRW}. 
\textit{Disentangled models}: \textbf{DR-CFR}~\citep{Hassanpour2020LearningDR}, \textbf{TEDVAE}~\citep{Zhang2021TreatmentEE}, \textbf{MIM-DRCFR} (our method) and its variant \textbf{RLO-DRCFR}.\par
\textbf{Ablation Study.} We also conduct an ablation study to examine the contributions of different components in MIM-DRCFR.\par

\begin{center}
\begin{table}[h]                  
\setlength{\abovecaptionskip}{0pt}%
\setlength{\belowcaptionskip}{0pt}%
\caption{Results of different treatment effect estimation methods on IHDP Benchmark and ablation study of MIM-DRCFR}
\centering
\begin{threeparttable}
\begin{tabular}{lllll}
\toprule[1pt]
\textsc{Dataset} & \multicolumn{2}{c}{IHDP-A}  & \multicolumn{2}{c}{IHDP-B} \\ \midrule
\textsc{Method} &
  \multicolumn{1}{c}{$\sqrt{\epsilon_{\textit{PEHE}}}$} &
  \multicolumn{1}{c}{$\epsilon_{\textit{ATE}}$} &
  \multicolumn{1}{c}{$\sqrt{\epsilon_{\textit{PEHE}}}$} &
  \multicolumn{1}{c}{$\epsilon_{\textit{ATE}}$}  \\ \midrule
\multicolumn{1}{l}{TARNET}  & 0.95 \footnotesize(0.38) & 0.27 \footnotesize(0.13)  & 3.15 \footnotesize(0.22)  & 0.42 \footnotesize(0.17) \\
\multicolumn{1}{l}{CFR-MMD}  & 0.75 \footnotesize(0.34) & 0.30 \footnotesize(0.12)   & 2.58 \footnotesize(0.18) & 0.35  \footnotesize(0.16) \\
\multicolumn{1}{l}{CFR-WASS} & 0.74 \footnotesize(0.35) & 0.29 \footnotesize(0.12)  & 2.51 \footnotesize(0.18) & 0.34   \footnotesize(0.16)\\
\multicolumn{1}{l}{CFR-ISW}   & 0.69 \footnotesize(0.30) & 0.23 \footnotesize(0.09) & 2.55 \footnotesize(0.16) & 0.40   \footnotesize(0.13) \\
\multicolumn{1}{l}{DR-CFR}    & 0.64 \footnotesize(0.25) & 0.20 \footnotesize(0.08) & 2.33 \footnotesize(0.15)& 0.37  \footnotesize(0.10) \\ 
\multicolumn{1}{l}{TEDVAE}    & 0.58 \footnotesize(0.22) & 0.15 \footnotesize(0.08) & 2.24 \footnotesize(0.13) & 0.31   \footnotesize(0.09) \\ \midrule
\multicolumn{1}{l}{RLO-DRCFR}  & 0.54 \footnotesize(0.16) & 0.14  \footnotesize(0.05) & 2.16 \footnotesize(0.11) & 0.31   \footnotesize(0.06)\\  
\multicolumn{1}{l}{MIM-DRCFR}  & \textBF{0.38 \footnotesize(0.09)} & \textBF{0.09 \footnotesize(0.01)} & \textBF{2.08 \footnotesize(0.09)} & \textBF{0.25 \footnotesize(0.04)} \\ \midrule \midrule
\multicolumn{1}{l}{\textit{w/o} SFD}  & 0.53 \footnotesize(0.20) & 0.14 \footnotesize(0.05) & 2.28 \footnotesize(0.13)& 0.34 \footnotesize(0.08)  \\
\multicolumn{1}{l}{\textit{w/o} MIM}   & 0.50 \footnotesize(0.21)& 0.13 \footnotesize(0.05) & 2.29 \footnotesize(0.12) & 0.32 \footnotesize(0.09)  \\
\multicolumn{1}{l}{\textit{w/o} Both} & 0.63 \footnotesize(0.25) & 0.19 \footnotesize(0.07) & 2.31 \footnotesize(0.15) & 0.37 \footnotesize(0.10) \\
\bottomrule[1pt]
\end{tabular}
\begin{tablenotes}
\footnotesize
\item[1] The \textbf{bolded} values mean the best performance and the metric represented in the form of ``mean (standard deviation)'' and the result is statistically significant based on the Welch’s $t$-test with $\alpha=0.05$.
\item[2] \textit{w/o} SFD means using three separate representation layers instead of SFD layer. \textit{w/o} Both means \textit{w/o} MIM+SFD.
\end{tablenotes}
\end{threeparttable}
\label{1}
\end{table}
\end{center}

In Table \ref{1}, we report the average results of the $\sqrt{\epsilon_{\textit{PEHE}}}$ and $\epsilon_{\textit{ATE}}$ metrics on IHDP-A and IHDP-B benchmarks (100 realizations with 63/27/10 proportion of train/validation/test splits). Results show that MIM-DRCFR achieves the best performance against the compared methods and its variants, which demonstrates that MIM-DRCFR is currently the most effective disentangled method in ITE estimation. The bottom part of Table \ref{1} summarizes results of the ablation study, which demonstrate that all MIM-DRCFR variants with some components removed witness clear performance drops when comparing to the full model on the $\sqrt{\epsilon_{\textit{PEHE}}}$ metric, indicating that each of the designed components contributes to the success of MIM-DRCFR.\par

\subsection{Real-world dataset Offline Evaluation}
In real-world scenarios we often face the following budget constrained problem \textit{“How to maximize global value of the population $\omega$ when we can only \textbf{intervene} on a subgroup $\lambda$ of the population $\omega$ due to the budget limit”}, which can be formulated as: 
\begin{equation}
\begin{split}
\max \sum_{i \in \lambda}y^{1}_{i}(x_{i},t_{i}&=1) + \sum_{i \in \delta}y^{0}_{i}(x_{i},t_{i}=0)\\
s.t.\sum_{i \in \lambda } \mathbb{I}[t_{i}&=1]\leq B,
\end{split}
\end{equation}
where $\lambda$ (resp., $\delta$) refers to treatment (resp., control) subgroup, $\omega = \lambda \cup \delta$ and $\lambda \cap \delta = \emptyset$. $\mathbb{I}$ denotes the indicator function and B refers to the budget (e.g., total number of treated users). We prove that this problem equals finding an optimal subgroup $\lambda^{*}$ that has higher non-negative \textit{uplift value} (i.e., individual treatment effect $\tau$) than that of $\delta$. Mathematically:
\begin{equation}
\begin{split}
\lambda^{*}=\left \{x_{i} \,| \,\forall x_{j} \in \delta, \, \tau_{i} \geqslant \tau_{j}, \, \tau_{i} \geqslant 0 \right \} \quad and \quad \lvert \lambda^{*} \rvert \leqslant  B
\end{split}
\end{equation}\par
We can easily obtain $\lambda^{*}$ through greedy approximation algorithm~\citep{Dantzig1957DiscreteVariableEP} based on uplift value. Thus, we convert this budget constrained problem into ITE estimation problem, which has gained lots of interest in recent years under the name of \textit{\textbf{uplift modelling}}. The problem consists in targeting treatment to the individuals for whom it would be the most beneficial. For instance, in marketing, one would aim to target advertisement budget to users that would be most likely to be persuadable to purchase ~\citep{betlei2020treatment}.\par
\textbf{Message pushing Dataset.} A real-world industrial dataset with 10 million samples that was collected from a current online policy.
The covariates contain users’ profile, behavior and context features, treatment is defined as \textit{``If pushing the message to user''} and the outcome is whether user log onto apps that day. In order to satisfy the \textit{unconfoundedness} assumption, we introduce sufficient confounders (e.g, user activity features) based on our prior knowledge.  \par
\textbf{Coupon sending Dataset.} This is similar to above with the message pushing action simply replaced by sending coupons.\par

\begin{table}[]
\setlength{\abovecaptionskip}{0pt}%
\setlength{\belowcaptionskip}{3pt}%
\caption{Offline AUUC on the two real-world datasets}
\begin{threeparttable}
\begin{tabular}{llll}
\toprule[1pt]
\multicolumn{1}{l}{\textsc{Method}} &
\multicolumn{1}{c}{Coupon sending} &
\multicolumn{1}{c}{Message pushing} \\ \midrule
\multicolumn{1}{l}{TARNET}   & \multicolumn{1}{c}{1.09} & \multicolumn{1}{c}{0.56} & \\
\multicolumn{1}{l}{DR-CFR}  & \multicolumn{1}{c}{1.22} & \multicolumn{1}{c}{0.68 } &  \\
\multicolumn{1}{l}{RLO-DRCFR}  & \multicolumn{1}{c}{1.41} & \multicolumn{1}{c}{0.73} & \\
\multicolumn{1}{l}{MIM-DRCFR}  & \multicolumn{1}{c}{\textBF{1.55}} & \multicolumn{1}{c}{\textBF{0.80}} \\
\bottomrule[1pt]
\end{tabular}
\begin{tablenotes}
\footnotesize
\item[1] We normalize the $AUUC$ value by dividing the $AUUC_{\pi}(1)$ .
\end{tablenotes}
\end{threeparttable}
\label{2}
\end{table}

\textbf{Performance Metrics.} As we cannot obtain factual and counterfactual outcomes simultaneously in the real-world datasets, we use Area Under the Uplift Curve (AUUC) \cite{gutierrez2017causal,zhang2021unified,Diemert2018ALS} as our offline metric:
\begin{equation}
AUUC_{\pi}(k)=(\frac{R^{T=1}_{\pi}(k)}{N^{T=1}_{\pi}(k)}-\frac{R^{T=0}_{\pi}(k)}{N^{T=0}_{\pi}(k)})(N^{T=1}_{\pi}(k)+N^{T=0}_{\pi}(k)),
\end{equation}
where $\pi(k)$ denotes the first k proportions of population sorted in descending order of uplift value and $k \in [0,1]$. $R^{T=1}_{\pi}(k)$ (resp., $R^{T=0}_{\pi}(k)$) are the positive outcomes (i.e., login outcome in our industrial datasets) in the treatment (resp., control) group and $N^{T=1}_{\pi}(k)$ (resp., $N^{T=0}_{\pi}(k)$) are the number of subjects in the treatment (resp., control) group from $\pi(k)$. The total AUUC is then obtained by cumulative summation~\citep{Diemert2018ALS}:
\begin{equation}
AUUC=\int_{0}^{1}AUUC_{\pi}(\rho)d\rho \approx \sum_{k=0}^{1}AUUC_{\pi}(k)dk
\end{equation}\par
\begin{figure}[h]
\setlength{\abovecaptionskip}{2pt}%
\centerline{\includegraphics[width=0.32\textwidth,height=0.22\textwidth]{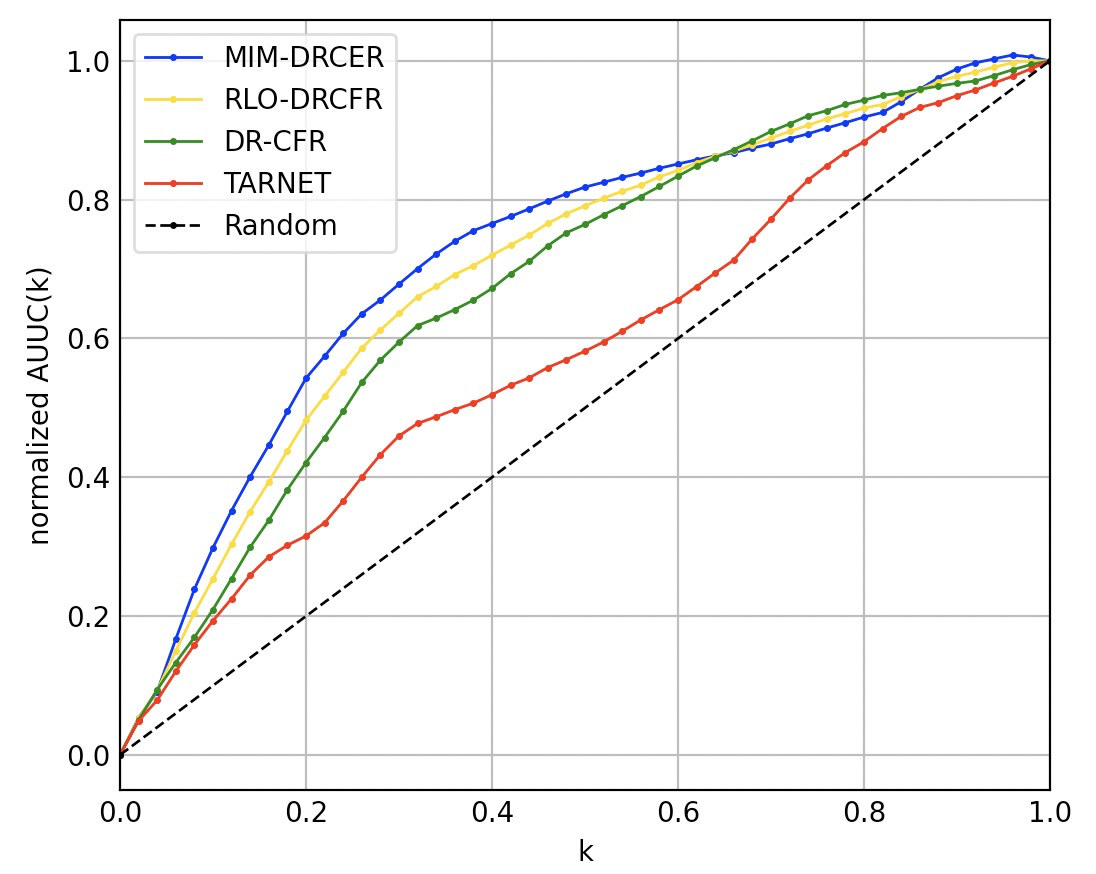}}
\caption{Uplift curve of the Messing pushing dataset. x-axis refers to the proportion $k$ of the test dataset, y-axis denotes the normalized $AUUC_{\pi}(k)$ value by dividing the $AUUC_{\pi}(1)$.}
\label{fig2}
\end{figure}

Table \ref{2} and Figure \ref{fig2} illustrate the offline AUUC and uplift curve, which demonstrates that MIM-DRCFR performs better than other methods on the real-world industrial datasets. \par

\subsection{Online A/B test}
We design an online A/B test to further evaluate the performance through calculating the daily login users (DLU) of the population $\omega$ after pushing message to the estimated optimal subgroup $\hat{\lambda}^{*}$.
\begin{equation}
    DLU(\omega) = L^{T=1}(\hat{\lambda}^{*}) + L^{T=0}(\omega - \hat{\lambda}^{*}),
\end{equation}
where $L^{T=1}(\hat{\lambda}^{*})$ (resp., $L^{T=0}(\omega - \hat{\lambda}^{*})$) refers to the number of login users of the treatment (resp., control) group and $\hat{\lambda}^{*}$ is selected based on the estimated uplift value of different models. In our online datasets, for each experiment group, $\omega$ contains 1 million users and these users are randomly sampled from the entire population who have logged onto our platform in the last 7 days. Finally we choose $60\%$ of them to push messages. \par
Table \ref{3} illustrates the online DLU result of different models during 5 days' experiment period. The average DLU result of MIM-DRCFR increased by \textbf{6.1\%} compared with current online policy, which is greater than that of DR-CFR and RLO-DRCFR. This again demonstrates the improvement of MIM-DRCFR on ITE estimation.

\begin{table}[h]
\setlength{\abovecaptionskip}{0pt}%
\setlength{\belowcaptionskip}{3pt}%
\caption{Comparison of DLU of Message Pushing}
\label{online_ab}
\setlength{\tabcolsep}{1.5mm}{
\begin{tabular}{lllllll}
\toprule[1pt]
\multicolumn{1}{l}{\textsc{Method}} &
\multicolumn{1}{c}{T} &
\multicolumn{1}{c}{T+1} &
\multicolumn{1}{c}{T+2} &
\multicolumn{1}{c}{T+3} &
\multicolumn{1}{c}{T+4} &
\multicolumn{1}{c}{Avg} \\ \midrule
\multicolumn{1}{l}{DR-CFR}     & \multicolumn{1}{c}{+2.1\%} & \multicolumn{1}{c}{+1.5\%} & \multicolumn{1}{c}{+1.2\%} & \multicolumn{1}{c}{+0.4\%} & \multicolumn{1}{c}{+0.8\%} & \multicolumn{1}{c}{+1.2\%}\\
\multicolumn{1}{l}{RLO-DRCFR}  & \multicolumn{1}{c}{+4.4\%} & \multicolumn{1}{c}{+4.8\%} & \multicolumn{1}{c}{+5.0\%} & \multicolumn{1}{c}{+3.5\%} & \multicolumn{1}{c}{+2.9\%} &\multicolumn{1}{c}{+4.1\%} \\
\multicolumn{1}{l}{MIM-DRCFR}  & \multicolumn{1}{c}{\textBF{+6.6\%}} & \multicolumn{1}{c}{\textBF{+8.5\%}} & \multicolumn{1}{c}{\textBF{+6.3\%}} & \multicolumn{1}{c}{\textBF{+5.8\%}} & \multicolumn{1}{c}{\textBF{+3.4\%}} &\multicolumn{1}{c}{\textBF{+6.1\%}}\\
\bottomrule[1pt]
\end{tabular}}
\label{3}
\end{table}

\section{Conclusion}
In this paper, we focus on disentangled representation learning for ITE estimation and propose a disentangled framework called MIM-DRCFR, which incorporates multi-task learning for the sake of information sharing during the disentangling process and MI minimization for obtaining better independence of the latent factors. Both public benchmarks and real-world industrial datasets demonstrate its superiority over state-of-the-art methods. For future work, we would like to explore more efficient disentangling framework like generative models and extend our method to multi-treatment scenarios.



\bibliographystyle{ACM-Reference-Format}
\balance
\bibliography{sample-sigconf.bbl}

\end{document}